\documentclass{article}

\usepackage[final,nonatbib]{nips_2017}

\usepackage[utf8]{inputenc} 
\usepackage[T1]{fontenc}    
\usepackage{hyperref}       
\usepackage{url}            
\usepackage{booktabs}       
\usepackage{amsfonts}       
\usepackage{nicefrac}       
\usepackage{microtype}      
\usepackage{wrapfig}

\usepackage{graphicx} 
\usepackage{epsfig} 

\usepackage[ruled,vlined,linesnumbered]{algorithm2e}

\usepackage{color}

\usepackage{hyperref}

\usepackage{graphicx} 
\usepackage{epsfig}
\usepackage{float}
\usepackage{caption}
\usepackage{subcaption}

\usepackage{amssymb}
\usepackage{amsmath}
\usepackage{amsthm}
\usepackage{mathtools}
\usepackage[raggedright]{titlesec} 

\def \mat     [#1]{\mathbf {#1}}
\def \matEqSub[#1 #2]{\mathbf{#1}_{#2}}
\def \matEquation[#1]{{\mathbf{#1}}}
\def \matCell[#1]{\mbox{$\bf #1_{cell}$}}
\def\Supper[#1 #2]{\mathcal{S}^{\rm upper}_{#1#2}}
\def\Slower[#1 #2]{\mathcal{S}^{\rm lower}_{#1#2}}

\def\Gupper[#1]{g^{\rm upper}_{#1}}
\def\Glower[#1]{g^{\rm lower}_{#1}}

\newcommand{\matrixdim}[2] {#1 \times #2}

\newcommand{\slower}{$\mathcal{S}^{\rm lower}$}
\newcommand{\supper}{$\mathcal{S}^{\rm upper}$}
\newcommand{\sstructure}{$\mathcal{S}^{\rm struct}$}

\newcommand{\fnormSqVanilla}[1]{\left\lVert#1\right\rVert_{F}^2}

\newcommand{\changeBM}[1]{{#1}}

\title{A two-dimensional decomposition approach for\\ matrix completion through gossip}

\author{
  Mukul Bhutani \\
  Amazon.com\\
  \texttt{mbhutani@amazon.com} \\
  \And
  Bamdev Mishra \\
  Amazon.com\\
  \texttt{bamdevm@amazon.com} \\
}

\begin{document}

\maketitle

\begin{abstract}
Factoring a matrix into two low rank matrices is at the heart of many problems. The problem of matrix completion especially uses it to decompose a sparse matrix into two non sparse, low rank matrices which can then be used to predict unknown entries of the original matrix. We present a scalable and decentralized approach in which instead of learning two factors for the original input matrix, we decompose the original matrix into a grid blocks, each of whose factors can be individually learned just by communicating (gossiping) with neighboring blocks. This eliminates any need for a central server. We show that our algorithm performs well on both synthetic and real datasets.   
\end{abstract}

\section{Introduction}

The problem of approximating a matrix by decomposing it into two low rank factor matrices is useful in solving many machine learning problems \cite{Azar.2001} including that of recommendation systems \cite{Drineas02}. The problem of building a recommendation system involves having a large, sparse matrix $\mat[X]$ of dimensions $\matrixdim{m}{n}$ and finding its low rank decomposition $\mat[X =  UW^{\top}]$ where $\mat[U]$ and $\mat[W]$ are non sparse matrices of dimensions $\matrixdim{m}{r}$ and $\matrixdim{n}{r}$ respectively and $r \ll m,n$. The product of these matrices can then be used to find the missing entries of $\mat[X]$. Since recommendation systems generally deal with user data, the security aspect may also become a major concern as one doesn't want to compromise the private data of users. 

The problem of matrix completion is treated as an optimization problem and can be solved using gradient search \cite{Gorrell, Guan}. Parallel versions of gradient search have been used but they still require a central server \cite{Chu, Le12}. This dependency on a central server during most part of learning is one which we intend to eliminate. \cite{Ling12} follows an approach in which the input matrix is decomposed (as groups of columns) into $l$ parts which were all processed by different agents. Each agent $i$ estimates its own version of the original matrix, $\matEqSub[X i]$, as $\matEqSub[UW i]^{\top}$. Also, the matrix $\mat[U]$ has to be synchronized between all the agents after each round of iteration. And thus each agent has the same (and also complete) view of $\mat[U]$ after each iteration. In the approach followed by \cite{Hegeds14}, in the decomposition $\mat[X = UW]^{\top}$, each row of $\mat[X]$ and $\mat[U]$ \changeBM{is} stored in different nodes and a public matrix $\mat[W]$ is exchanged between them. Random walks are performed to bring convergence in $\mat[W]$. However, here too a single agent takes care of the complete row. 

In contrast to the works discussed above, we present a decomposition strategy in a way such that the matrix is decomposed not only row wise \cite{Mishra16}, but also in a column wise fashion. We decompose the matrix $\mat[X]$ into a two dimensional rectangular grid such that each decomposed part (which we henceforth call a block) can be factorized into its own local, row rank factors $\mat[U]$s and $\mat[W]$s. These blocks can then be processed by different agents. The individual blocks just {\it gossip} with their neighbors, trying to reach consensus and thus no communication happens with a central server during the learning phase. Once the learning is done, a final culmination of $\mat[U]$s and $\mat[W]$s is performed. 

\section{Decomposition pattern}\label{sec:decomposition}
\begin{figure}[t]
    \centering
           \includegraphics[width=0.7\textwidth]{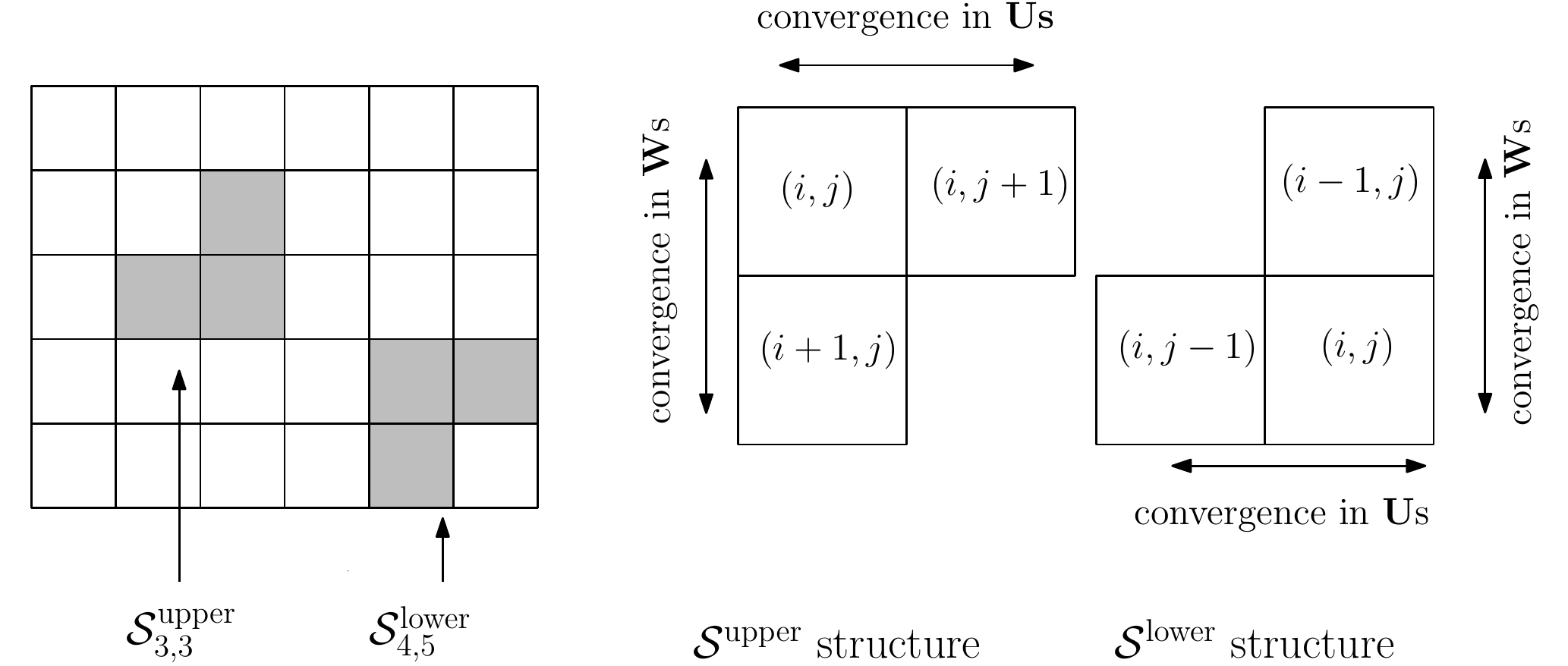}
             \caption{General decomposition of a matrix $\mat[X]$ into a grid of $\matrixdim{5}{6}$ blocks. Two particular structures, $\Supper[4 5]$ and $\Slower[3 3]$, are highlighted. If $\mat[X]$ had dimensions $\matrixdim{500}{600}$, then each of the $\matrixdim{5}{6}$ block would have $\matrixdim{100}{100}$ entries.}
    \label{fig:decomposition}
\end{figure}

The input matrix $\mat[X]$ is decomposed into a $\matrixdim{p}{q}$ dimensional rectangular grid of blocks (Figure \ref{fig:decomposition}). Each block can be referred by using the indices $i$ and $j$ corresponding to its row and column in the decomposition respectively. Each $\matEqSub[X ij]$ can then be factored as $\matEqSub[U ij]$ $\matEqSub[W^{\top} ij]$ as usual. We try to learn $\mat[U]$s and $\mat[W]$s corresponding to each of these blocks and then in the end, combine them appropriately to form universal $\mat[U]$ and $\mat[W]$ which can then be used to find the missing entries of the original matrix $\mat[X]$. These individual blocks gossip with their neighbors (blocks sharing an edge) and try to reach a consensus. Each row tries to reach to a consensus in terms of $\mat[U]$ and each column tries to reach a consensus for $\mat[W]$. All these $\mat[U]$s and $\mat[W]$s are appended together to from the universal $\mat[U]$ and $\mat[W]$ which represents factors of original matrix $\mat[X]$.
This pattern of communication results in a natural structure of a group of blocks, which can be thought of as gossiping. Let's call these groups of blocks or structures as \supper and \slower (shown in Figure \ref{fig:decomposition}). We call one of the blocks of the structure as pivot block indexed as $(i,j)$ and the other two blocks are indexed relatively. Each block can belong to one or more structures depending on its position in the grid. Each of these structures is an independent computational unit and thus any two non overlapping structure can potentially be processed independently by different agents. 

\section{Problem formulation} 
\label{sec:problemFormulation}
We model the matrix completion problem as an optimization problem and the objective function of our two dimensional decomposed formulation can be derived by doing the analysis of our \supper and \slower structures. For \supper, for blocks $(i, j)$ and $(i+1, j)$, we try to bring convergence between their $\mat[W]$s and for the blocks $(i, j)$ and $(i, j+1)$ we try to bring the convergence to their $\mat[U]$s. We define the cost of a structure as comprising of two components: $f$ and $d$. 
The $f$ cost component of a block measures how close it is to the original matrix and the $d$ cost component measures the consensus between two adjacent $\mat[U]$s (denoted as $d^{U}$) or $\mat[W]$s (denoted as $d^{W}$). For a block indexed at $(i, j)$ it can be written as:
\begin{equation}\label{eq:dcostu} 
\begin{array}{ll}
f_{ij} = \fnormSqVanilla{\matEqSub[X ij] - \matEqSub[U ij]\matEqSub[W^{\top} ij]},
d_{ij}^{U} = \fnormSqVanilla{\matEqSub[U ij] - \matEqSub[U ij+1]}, \text{ and }
  d_{ij}^{W} = \fnormSqVanilla{\matEqSub[W ij] - \matEqSub[W i+1j]},
\end{array}
\end{equation}
where $\fnormSqVanilla{\mat[Z]}$ denotes the square of the Frobenius norm of $\mat[Z]$. Consequently, the the total cost ($g$) for a structure turns out to be:
\begin{equation*}
g^{{\rm structure}} = f_{\text{(for all the three blocks)}} + \rho d^{U} + \rho d^{W},
\end{equation*}
where $\rho$ is the {\it weight} factor. Hence, the total cost for an \supper structure turns out to be:
\begin{equation} \label{eq:supper_cost}
\Gupper[ij] = f_{ij} + f_{i+1j} + f_{ij+1} + \rho \fnormSqVanilla{\matEqSub[U ij] - \matEqSub[U ij+1]}  + \rho \fnormSqVanilla{\matEqSub[W ij] - \matEqSub[W i+1j]}.
\end{equation}
For \slower, we can derive the costs in similar fashion. For a decomposition of matrix $\mat[X]$ into $\matrixdim{p}{q}$, our end goal is to minimize the sum of costs for all \supper and \slower possible, i.e.,
\begin{equation}
\label{eq:finalObjectiveFunction}
\min_{\matEqSub[U ij], \matEqSub[W ij]} \sum_{i=1, j=1}^{p, q} \Gupper[ij] + \Glower[ij] + \lambda\fnormSqVanilla{\matEqSub[U ij]} + \lambda\fnormSqVanilla{\matEqSub[W ij]},
\end{equation}
where the cost for $\Gupper[ij]$ (and similarly $\Glower[ij]$) can be seen from (\ref{eq:supper_cost}) if a \sstructure\ is valid or is 0 otherwise and $\lambda$ is the regularization \changeBM{parameter} added according to \cite{Bottou17}. 

\section{Algorithm}
\label{sec:algorithm}
Our basic online sequential algorithm in Algorithm \ref{OnlineAlgo} is very simple and uses the stochastic gradient descent (SGD) algorithm for (\ref{eq:finalObjectiveFunction}). We have an input matrix $\mat[X]$ of dimensions $\matrixdim{m}{n}$, which we divide into a grid of $\matrixdim{p}{q}$ blocks. Here $p$ and $q$ are the hyper parameters which would govern how one wishes to distribute the data. This in turn can depend upon the number of agents which one wishes to employ. Each of these blocks can be factored into corresponding $\matEqSub[U ij]$ and $\matEqSub[W ij]$ having rank $r$. These $\mat[U]$s and $\mat[W]$s are initialized randomly. We then randomly sample a structure out of the various possible ones (details of which were described in Section \ref{sec:decomposition}), $\rm{updateThroughSGD}$ calculates the gradient and updates the corresponding three $\mat[U]$s and $\mat[W]$s (corresponding to the three blocks comprising that structure). The process of sampling and updating parameters is repeated until convergence is reached. To update the $\lambda$ \cite{Bottou17} we use $\gamma_{t} = a/(1 + (bt))$, where $t$ is the number of iterations and $a,b$ are scalars. After Algorithm \ref{OnlineAlgo} has converged, all the $\mat[U]$s and $\mat[W]$s are finally combined to form $\mat[U]$ and $\mat[W]$ of size $m\times r$ and $n\times r$, respectively.

\begin{figure}[t]
    \begin{subfigure}[t]{0.3\textwidth}
    \centering
        \includegraphics[width=0.45\linewidth]{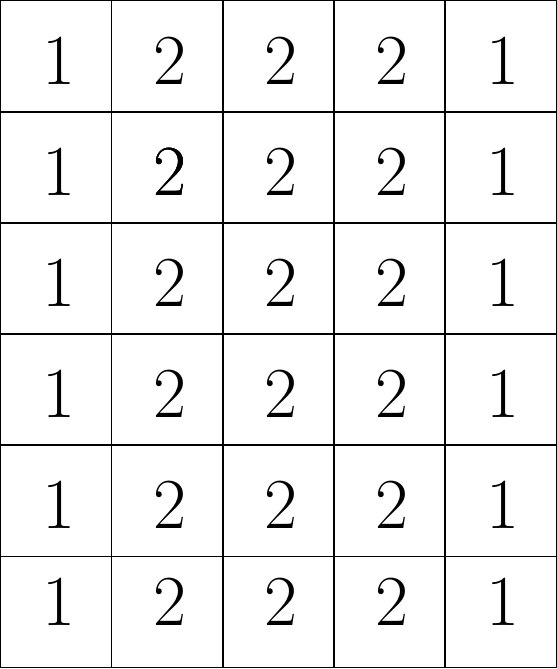}
        \caption{The relative number of times a block is selected while calculating the gradient of $d^{U}$.}
        \label{fig:dcostU}
    \end{subfigure}\hfill%
    \begin{subfigure}[t]{0.3\textwidth}
    \centering
        \includegraphics[width=0.45\linewidth]{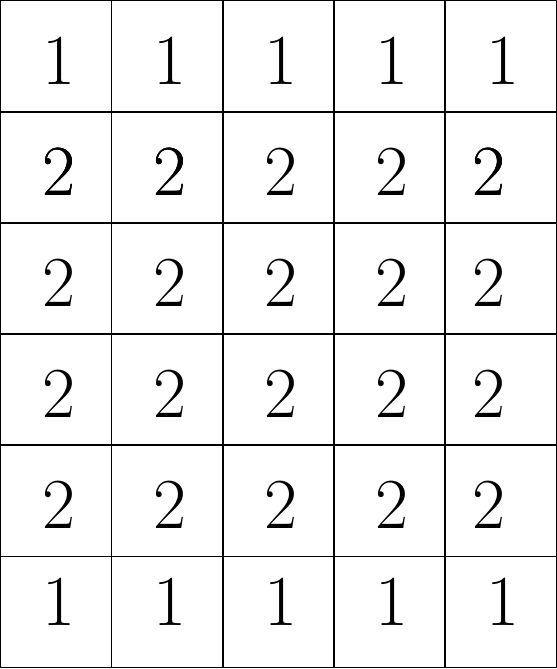}
        \caption{The relative number of times a block is selected while calculating the gradient of $d^{W}$.}
        \label{fig:dcostW}
    \end{subfigure}\hfill%
    \begin{subfigure}[t]{0.3\textwidth}
    \centering
        \includegraphics[width=0.45\linewidth]{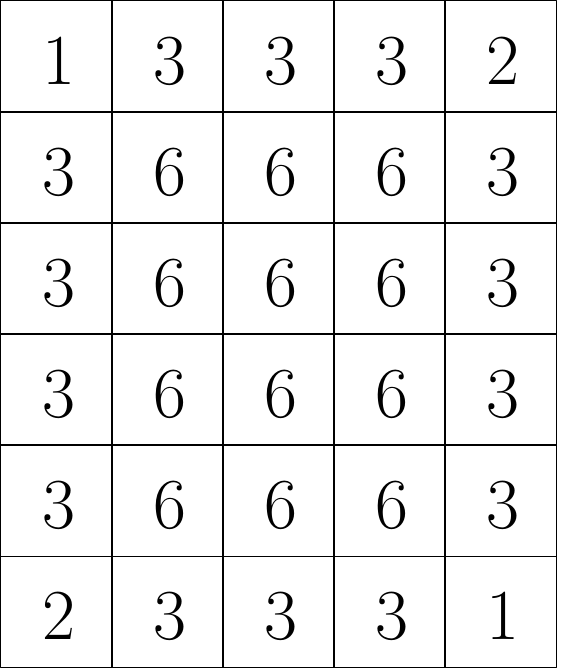}
        \caption{The number of times a particular block gets repeated while calculating gradient for $f$.}
        \label{fig:fcost}
    \end{subfigure}\hfill%
    \caption{Relative frequencies of selection of different blocks for a grid of size $6\times 5$.}
    \label{fig:normalization}
\end{figure}

\textbf{Normalizing representations of blocks.} Owning to our methodology of decomposition, the number of structures of which a block may be a part of is different for various blocks. Since, finally in (\ref{eq:finalObjectiveFunction}) we want all the blocks to have equal representation, we multiply a coefficient to each block to normalize the times a block may be selected for an update. The relative frequency of a block getting selected in given in Figure \ref{fig:normalization} and thus the coefficients we use are the inverse of it. 
\begin{algorithm}[h]
\scriptsize
\SetKwData{Left}{left}\SetKwData{This}{this}\SetKwData{Up}{up}
\SetKwFunction{Union}{Union}\SetKwFunction{FindCompress}{FindCompress}
\SetKwInOut{Input}{input}\SetKwInOut{Output}{output}
\DontPrintSemicolon
\SetAlgoLined
\Input{Decomposed blocks for $\mat[X]$ and rank $r$.}
\Output{$\mat[U]$s, $\mat[W]$s.}
\BlankLine
Initialize all $\mat[U]$s and $\mat[W]$s.\;

\While{\text{convergence is not reached}}{
    \sstructure $=$ randomly pick a valid structure.\;
    [$\mat[U]$s, $\mat[W]$s] = ${\rm updateThroughSGD}$($\mat[X]$s, \sstructure).\;
    Check for convergence.
} 
\caption{Basic update algorithm via SGD}
\label{OnlineAlgo}
\end{algorithm} 

\section{Numerical experiments}
\label{sec:experimentation}
To demonstrate the efficacy of our algorithm, we perform two sets of experiments. In the first set, we run our algorithm on synthetically generated data sets and calculate the cost. For the second set of experiments, we demonstrate our algorithm working on some popular public datasets. We use the root mean squared error (RMSE) to gauge the performance of our algorithm. 

\textbf{Experiments on synthetic data sets.} We randomly generate a synthetic matrix subject to a rank constraint. Of this we mask majority of the elements thus generating our train matrix. Similarly, a test matrix is also generated by choosing a fraction of elements in the original matrix which got masked and weren't selected for generating the train matrix and is used for evaluation. 


Table \ref{table:syntheticParams} describe various hyper parameters which were used for the experiments and Table \ref{table:syntheticExp} shows the costs (i.e., $\sum_{i=1, j=1}^{p, q} f_{ij} + \lambda\fnormSqVanilla{\matEqSub[U ij]} + \lambda\fnormSqVanilla{\matEqSub[W ij]}$) as iterations increase. 

\begin{table}[t]
\caption{Parameters used for various experiments.} 
\scriptsize
\centering 
\begin{tabular}{c c c c c c c c}
\toprule
Parameter & Exp\#1 & Exp\#2 & Exp\#3 & Exp\#4 & Exp\#5 & Exp\#6 \\
\midrule
$\rho$ (weight factor)   & 1e3     & 1e3         & 1e3       & 1e3       & 1e3       &  1e3\\
$\lambda$  (regularization parameter) & 1e-9     & 1e-9        & 1e-9  & 1e-9  & 1e-9  &  1e-9\\
$\matrixdim{p}{q}$ (dimensions of decomposed grid) & $\matrixdim{4}{4}$ & $\matrixdim{4}{5}$ & $\matrixdim{5}{5}$ & $\matrixdim{6}{6}$ & $\matrixdim{5}{5}$ & $\matrixdim{5}{5}$\\
$\matrixdim{m}{n}$ (input matrix dimensions) & $\matrixdim{500}{500}$ & $\matrixdim{500}{500}$ & $\matrixdim{500}{500}$ & $\matrixdim{500}{500}$ & $\matrixdim{5000}{5000}$ & $\matrixdim{10000}{10000}$\\
$a$    (scalar in stepsize tuning)             & 5.0e-04 & 5.0e-04 & 5.0e-04 & 5.0e-04  & 5.0e-04 & 5.0e-04\\
$b$    (scalar in stepsize tuning)             & 5.0e-07 & 5.0e-07 & 5.0-07 & 5.0e-07  & 5.0e-06 & 5.0e-07\\
\bottomrule
\end{tabular}
\label{table:syntheticParams}
\end{table}

\begin{table}[t]
\caption{Empirical proof of convergence of the algorithm.}
\scriptsize
\centering 
\begin{tabular}{c c c c c c c c}
\toprule
NumIterations & Exp\#1 & Exp\#2         & Exp\#3        & Exp\#4        & Exp\#5        & Exp\#6\\
\midrule
0       & 1.45e+05    & 1.45e+05        & 1.45e+05      & 1.44e+05      & 6.42e+05      & 6.66e+07  \\
80000   & 6.92e-03    & 1.32e-01        & 1.45e+00      & 4.74e+02      & 1.26e+05      & 2.13e+04  \\
160000  & 9.62e-06    & 7.65e-05        & 1.44e-04      & 9.94e-01      & 2.83e+02      & 4.06e+00  \\
240000  & convergence & 1.07e-05        & 1.25e-05      & 1.04e-02      & 2.85e-01      & 9.96e-03  \\
260000  &             & convergence     & 1.21e-05      & 4.41e-03      & 7.39e-02      & 2.78e-03  \\
280000  &             &                 & convergence   & 1.96e-03      & 2.09e-02      & convergence \\
300000  &             &                 &               & 9.28e-04      & 6.44e-03      &                \\
400000  &             &                 &               & convergence   & convergence   &  \\
[1ex] 
\bottomrule
\end{tabular}
\label{table:syntheticExp} 
\end{table}

\begin{wraptable}{r}{7.5cm}
\caption{Experiments using real datasets.} 
\scriptsize
\begin{tabular}{c  c c c c c}
\toprule
\multicolumn{6}{c}{Number of blocks $\matrixdim{p}{q}$} \\
\midrule
Rank    & $\matrixdim{2}{2}$ &  $\matrixdim{3}{3}$  &   $\matrixdim{4}{4}$  & $\matrixdim{5}{5}$ &  $\matrixdim{10}{10}$ \\
\midrule
\multicolumn{6}{c}{MovieLens 1 million } \\
5   &   0.87    &   0.99    &   1.04    &   0.99    &   1.13 \\
10  &   0.86    &   0.99    &   1.03    &   1.00    &   1.22 \\
15  &   0.86    &   0.99    &   1.03    &   0.99    &   1.34 \\
\midrule
\multicolumn{6}{c}{\changeBM{MovieLens} 10 million } \\
5   &   0.97    &   0.95    &   0.98    &   0.97   &   1.07 \\
10  &   0.97    &   0.95    &   1.00    &   0.99    &   1.25 \\
15  &   0.98    &   0.96    &   1.03    &   1.02121     &   1.41\\
\midrule
\multicolumn{6}{c}{MovieLens 20 million } \\
5   &   0.95    &   0.92    &   0.93    &   0.99    &   1.01 \\
10  &   0.96    &   0.93    &   0.93    &   1.02    &   1.11 \\
15  &   0.96    &   0.94    &   0.93    &   1.05    &   1.24 \\
\midrule
\multicolumn{6}{c}{Netflix } \\
5    &   1.03   &  0.98 	 &  1.13  &  1.06  &  1.02 \\
10  &  1.00    &  0.98   &  1.14  &  1.02  &  1.02 \\
15  &   1.00   &  1.11    &  1.16 &  1.02  &  1.03 \\
\bottomrule
\end{tabular}\label{table:realDataset}
\end{wraptable} 

\textbf{Experiments on real data sets.} We demonstrate the efficacy of our algorithm on widely used, high dimensional, and highly sparse public datasets which are frequently used for benchmarking. The input data is split in a 80 - 20 ratio. The training is done with the 80 percent part. The 20 percent part is kept for testing on which we calculate and report the RMSE. All the experiments are performed with tuned parameters. Table \ref{table:realDataset} lists the RMSE we received for various datasets with different decomposition pattern ($\matrixdim{p}{q}$).

\textbf{Conclusion from the experiments.} As we can see from the above experiments, the algorithm is able to learn different $\mat[U]$s and $\mat[W]$s corresponding to different blocks and reaches to convergence in all the cases. The order of reduction of the cost on synthetic datasets is 7 to 10 in all the cases. The experimentation on real datasets also provide enough evidence on learning proving that we can learn global factors even though we may be processing the data in many smaller and independent parts.

\section{Conclusion}
We proposed a novel algorithm for matrix completion, where the data is decomposed along two dimensions, both row wise and column wise. It uses the gossip paradigm for communicating between the independent decomposed units. We get a decomposition in which individual decomposed units reach convergence by just talking to their neighbors, and hence, being independent of the master during the learning phase. Our initial experiments on the synthetic and real datasets shows the efficacy of our algorithm. Exploiting the fact that many of the \sstructure\ do not contain any overlapping blocks, and hence can be processed in parallel, will be a topic of future research.


\small
\bibliographystyle{unsrt}
\bibliography{myreferences}

\end{document}